\pdfoutput=1

\documentclass[11pt]{article}

\usepackage[preprint]{acl}

\usepackage{times}
\usepackage{latexsym}

\usepackage[T1]{fontenc}

\usepackage[utf8]{inputenc}

\usepackage{microtype}

\usepackage{inconsolata}

\usepackage{graphicx}
\usepackage{pifont}
\usepackage{booktabs}
\usepackage{subcaption}
\usepackage{algorithm}
\usepackage{algpseudocode}
\usepackage[most]{tcolorbox}
\usepackage[table, dvipsnames]{xcolor}
\usepackage{tabularx}
\usepackage[super]{nth}
\newcolumntype{Y}{>{\centering\arraybackslash}X}

\usepackage{xspace}

\def\Think{\textsc{Think}\xspace}
\def\Verbalize{\textsc{Verbalize}\xspace}
\def\Speak{\textsc{Speak}\xspace}

\def\model{\textsc{ReVerT}\xspace}
\def\framework{\textsc{Think-Verbalize-Speak}\xspace}
\def\SEQ{\textsc{Seq}\xspace}

\def\bov{$\langle\texttt{bov}\rangle$\xspace}
\def\eov{$\langle\texttt{eov}\rangle$\xspace}
\def\con{$\langle\texttt{con}\rangle$\xspace}

\def\meos{\langle\texttt{eos}\rangle\xspace}
\def\mbov{\langle\texttt{bov}\rangle\xspace}
\def\meov{\langle\texttt{eov}\rangle\xspace}
\def\mcon{\langle\texttt{con}\rangle\xspace}

\def\qwen{Qwen2.5-7B-Instruct\xspace}
\def\llama{Llama-3.1-8B-Instruct\xspace}
\def\gpt{gpt-4o-mini-2024-07-18\xspace}
\def\gptf{gpt-4.1-mini-2025-04-11\xspace}

\def\qwenv{Qwen2.5-3B-Instruct\xspace}
\def\qweno{Qwen2.5-Omni-7B\xspace}

\def\gsm8k{GSM8K\xspace}
\def\2wiki{2WikiMultiHopQA\xspace}
\def\mhqa{2MHQA\xspace}
\def\scibench{SciBench\xspace}

\usepackage{amsmath,amsfonts,bm}
\usepackage[dvipsnames]{xcolor}

\newcommand{\mstd}[2]{#1\textsuperscript{\tiny$\pm$#2}}
\newcommand{\bmstd}[2]{\textbf{#1}\textsuperscript{\tiny$\pm$#2}}
\newcommand{\umstd}[2]{\underline{#1}\textsuperscript{\tiny$\pm$#2}}

\newcommand{\correct}[1]{\textcolor{Green}{\textbf{#1}}}
\newcommand{\wrong}[1]{\textcolor{Red}{\textbf{#1}}}

\usepackage{xcolor}      
\usepackage{ifthen}      
\usepackage{environ}     

\newboolean{debug}

\NewEnviron{overview}{%
  \ifthenelse{\boolean{debug}}{%
    \textcolor{orange}{\BODY}%
  }{}%
}

\def\ie{\textit{i.e.}}

\def\ie{\emph{i.e.}}

\def\eqref#1{equation~\ref{#1}}

\def\1{\bm{1}}

\DeclareMathAlphabet{\mathsfit}{\encodingdefault}{\sfdefault}{m}{sl}
\SetMathAlphabet{\mathsfit}{bold}{\encodingdefault}{\sfdefault}{bx}{n}

\def\gC{{\mathcal{C}}}
\def\gD{{\mathcal{D}}}

\def\gQ{{\mathcal{Q}}}
\def\gR{{\mathcal{R}}}
\def\gS{{\mathcal{S}}}

\def\gV{{\mathcal{V}}}

\def\gX{{\mathcal{X}}}

\title{Think, Verbalize, then Speak: \\ Bridging Complex Thoughts and Comprehensible Speech}

\author{
    \textbf{Sang Hoon Woo}\thanks{Equal contribution.} \quad
    \textbf{Sehun Lee}\footnotemark[1] \quad
    \textbf{Kang-wook Kim} \quad
    \textbf{Gunhee Kim} \\
    Seoul National University \\
    \small{
        \texttt{tonyswoo@gmail.com}\quad
        \texttt{shlee@vision.snu.ac.kr}\quad
        \texttt{full324@snu.ac.kr}\quad
        \texttt{gunhee@snu.ac.kr}
    } \\
}

\makeatletter
\DeclareRobustCommand\onedot{\futurelet\@let@token\@onedot}
\def\@onedot{\ifx\@let@token.\else.\null\fi\xspace}

\makeatother

\usepackage{lipsum}
\usepackage{multirow}
\usepackage{multicol}
\usepackage{listings}
\setboolean{debug}{false} 

\begin{document}
\maketitle
\begin{abstract}
Spoken dialogue systems increasingly employ large language models (LLMs) to leverage their advanced reasoning capabilities. However, direct application of LLMs in spoken communication often yield suboptimal results due to mismatches between optimal textual and verbal delivery. While existing approaches adapt LLMs to produce speech-friendly outputs, their impact on reasoning performance remains underexplored. In this work, we propose \textbf{\framework}, a framework that decouples reasoning from spoken delivery to preserve the full reasoning capacity of LLMs. Central to our method is \textit{verbalizing}, an intermediate step that translates thoughts into natural, speech-ready text. We also introduce \textbf{\model{}}, a latency-efficient verbalizer based on incremental and asynchronous summarization. Experiments across multiple benchmarks show that our method enhances speech naturalness and conciseness with minimal impact on reasoning. The project page with the dataset and the source code is available at \url{https://yhytoto12.github.io/TVS-ReVerT}

\end{abstract}

\section{Introduction}

Humans inherently distinguish between their internal thoughts and their external expressions, effortlessly reformulating their thought processes into formats suitable for verbal communication~\citep{levelt1993speaking, indefrey2004spatial, sahin2009sequential}. Current spoken dialogue systems, despite rapid advances, lack mechanisms that emulate this fundamental human capacity. This limitation becomes increasingly significant as reasoning models that produce extensive chain-of-thought to address complex problems gain popularity~\citep{cot,o1,guo2025deepseek}.

\begin{figure}[tb!]
    \centering
    \includegraphics[width=\linewidth]{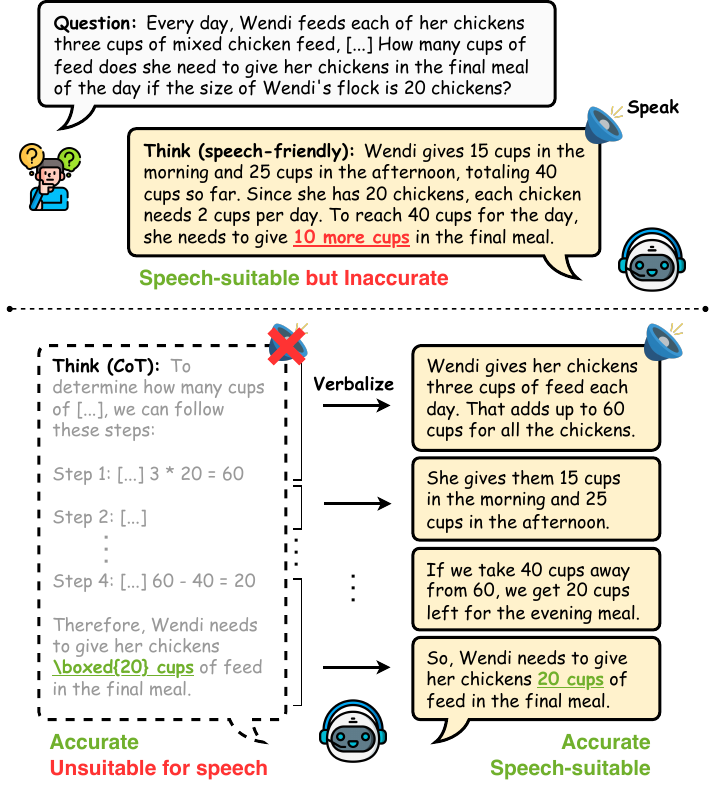}
    \caption{
    To produce both speech-friendly and accurate responses, we decouple thinking from verbalizing. A chain-of-thought process (verbose, structured, or in technical formats such as LaTeX) is unsuitable for spoken delivery. Conversely, generating a speech-friendly answer without underlying reasoning may be fast but often results in inaccurate responses. Moreover, waiting for the thinking to complete leads to severe latency. By incrementally verbalizing internal thoughts, we achieve reasoning capability, speech-suitability, and low latency.
    }
    \label{fig:intro}
    \vspace{-0.4cm}
\end{figure}

Current spoken dialogue systems typically employ a two-stage framework, herein referred to as the \textsc{Think-Speak} framework~\citep{wavchat, dongre2024respact, qwen2omni,fang2025llamaomni}. In this approach, the system first constructs the content of the speech (\Think), and then generates the corresponding spoken output (\Speak). However, large language models (LLMs), which are commonly used in the \Think stage, combined with test-time computing methods such as chain-of-thought reasoning, often yield responses that are not suitable for spoken dialogue. While some studies~\citep{speechworthy, mathreader} address this issue by guiding the model to produce speech-friendly outputs through fine-tuning or prompting, enforcing a speech-friendly thought formats may substantially deteriorate the reasoning performance. Figure \ref{fig:intro} illustrates an example of this issue.

We propose the \framework{} framework, which introduces an intermediate verbalization stage that translates raw model reasoning into speech-friendly, comprehensible utterances. Through this verbalization process, our system produces natural, concise speech output without sacrificing problem-solving capabilities. To mitigate the latency in the naive two-stage sequential implementation, we present the \textbf{RE}asoning to \textbf{VER}bal \textbf{T}ext (\model{}) model, which utilizes efficient, incremental verbalization and achieves up to 66\% reduction in response time compared to the sequential approach. Extensive automatic and human evaluations confirm that our method generates speech output that is both natural and accurate, with minimal loss in reasoning capabilities and robust performance across different reasoning models and verbalize model sizes.

Our key contributions to the field are as follows:
\begin{itemize}
    \item We introduce the \framework framework, which enhances the speech-friendliness of generated utterances while preserving the problem-solving capabilities of the underlying reasoning model.
    \item We propose \model, a latency-efficient verbalization model that significantly reduces system latency by performing verbalization in parallel with the underlying reasoning process.
    \item We develop the solve-summarize-scatter data pipeline that transforms existing question answering (QA) datasets into \model training datasets by generating reasoning sequences with incremental, speech-friendly summaries. We publicly release the dataset.
\end{itemize}

\section{Related Work} \label{sec:related_work}
\paragraph{Reasoning in LLMs}
While LLMs have achieved significant progress through model and dataset scaling, these advancements alone remain insufficient for addressing complex tasks such as arithmetic and commonsense reasoning~\citep{gsm8k,2wikimultihop,scibench,wang2024mmlu}. The introduction of chain-of-thought (CoT) prompting~\citep{cot} has unlocked enhanced reasoning abilities in LLMs.
Subsequent research has developed specialized reasoning models that incorporate non-linear reasoning processes, such as reflection and backtracking~\cite{o1, guo2025deepseek}. However, these enhanced reasoning processes are lengthy and verbose, making them difficult for users to stay engaged during spoken interactions.

\begin{figure*}[!tb]
\centering
\includegraphics[width=\linewidth]{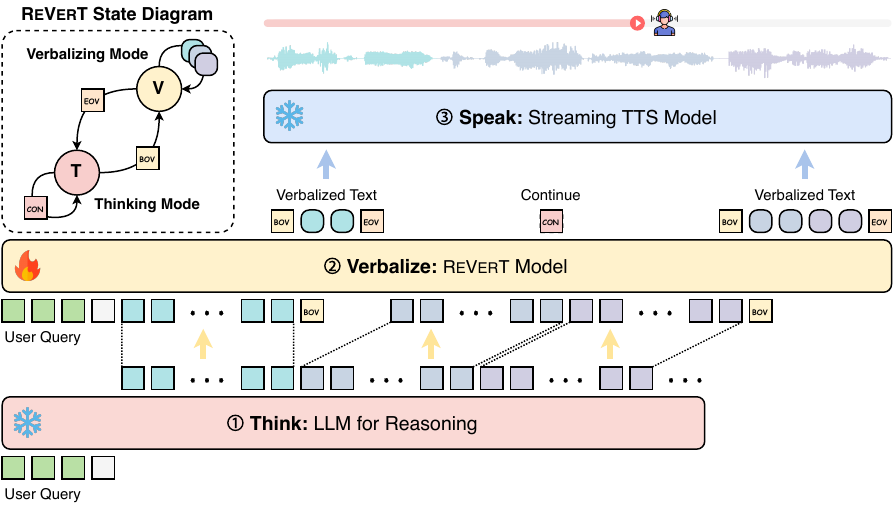}
\vspace{-0.7cm}
\caption{Overall framework of \framework. For a given user query, (1) a reasoning LLM generates a step-by-step chain-of-thought reasoning in text, (2) the \model{} model verbalizes the intermediate reasoning outputs into speech-friendly text incrementally to reduce the latency, and (3) a TTS model converts the verbalized text into synthesized speech output in a streaming manner. \model{} model operates in two modes: \textit{thinking mode} ($S_{T}$), where it receives and accumulates reasoning chunks, and \textit{verbalizing mode} ($S_{V}$), where it translates accumulated reasoning into speech-friendly text. Please refer to \S~\ref{sec:verbalize} for details regarding the usage of special tokens \bov{}, \con{}, and \eov{}.}
\vspace{-0.2cm}
\label{fig:architecture}
\end{figure*}

\paragraph{Spoken Dialogue Systems}
Spoken dialogue systems are typically categorized as cascaded or end-to-end~\citep{wavchat}. Cascaded systems employ a pipeline architecture comprising automatic speech recognition (ASR), a dialogue model, and a text-to-speech (TTS) component, using text as the intermediate representation. This modular approach allows for the integration of state-of-the-art components at each stage. However, LLM-based dialogue models within these systems often produce outputs optimized for reading, such as bullet points, sentence fragments, or formatted equations, rather than for spoken communication, which can undermine the naturalness of speech-based interactions.

End-to-end systems eliminate the dependency on intermediate text, thereby preserving paralinguistic cues and facilitating more natural speech generation. Recent work includes fully textless models~\citep{gslm,speechgpt,moshi}, text-speech interleaved architectures~\citep{glm4voice,freezeomni}, and parallel decoding approaches~\citep{miniomni,lucy,qwen2omni}. While end-to-end systems are more effective at generating speech-friendly outputs, they typically exhibit weaker reasoning capabilities compared to conventional LLMs.

\paragraph{Speech-Suitable Text Generation}
Recent work on speech-suitable text can be divided into two main categories. The first is normalization, which converts non-standard text into standard, pronounceable forms. For example, MathReader~\citep{mathreader} translates LaTeX mathematical expressions into English, which is crucial for LLM-based spoken dialogue systems since LLMs often output LaTeX equations when solving arithmetic problems.
The second category considers how content should be verbalized for effective spoken communication, based on the fundamental differences between textual and audio media. For instance, unlike text, audio requires listeners to engage with content sequentially, without the ability to selectively skip or return to sections. Building on this observation, \citet{speechworthy} introduces the concept of "speechworthiness," referring to properties that make text well-suited for verbal communication, including clarity, utterance length, and information density.

\section{\framework} \label{sec:approach}
Our framework, \framework, modifies the traditional cascaded system by generating response content in two stages: a reasoning stage that ensures response accuracy (\Think) and a translation stage that converts the reasoning output into a verbal response (\Verbalize). The system subsequently converts the resulting response to speech (\Speak). Figure \ref{fig:architecture} provides an overview of our approach with \model as the verbalizer. We employ an off-the-shelf reasoning LLM and a streaming TTS model, both of which remain frozen; only the \model model undergoes training under our framework.

\subsection{\textsc{Think}}

In the \Think stage, we leverage the problem-solving abilities of a reasoning LLM.
Upon receiving a user query, the LLM solves the query using chain-of-thought reasoning. The reasoning output is then streamed to the subsequent stage.

\subsection{\textsc{Verbalize}} \label{sec:verbalize}

In the \Verbalize stage, the system receives the streaming reasoning output from the \Think stage and translates it into speech-friendly utterances. A naive approach would be the sequential approach, where the system completes the reasoning stage and then generates the speech-friendly translations based on the complete output. However, the sequential approach introduces significant latency.

To address this issue, we propose \model, a latency-efficient verbalizer. As described in Algorithm~\ref{alg:revert}, the \model model operates asynchronously with the reasoning LLM from the \Think stage, incrementally generating speech-friendly utterances based on partial reasoning outputs.

The \model model operates in two distinct modes: \textit{thinking mode} ($\gS_{T}$) and \textit{verbalizing mode} ($\gS_{V}$). In thinking mode, \model receives and processes the outputs of the reasoning model. While the reasoning LLM emits output token by token, \model processes these outputs in segments, defined by a predetermined set of delimiters. This chunk-based processing enables more efficient computation through hardware parallelism.

After processing each segment, \model determines whether to initiate verbalization via single token generation. If the next token is \con, \model continues processing additional reasoning segments. If the next token is \bov, the model transitions to verbalizing mode, where \model translates the accumulated reasoning segments into speech-friendly output tokens. The model continues generating verbalized text until it produces the \eov token, at which point it forwards the generated text to the subsequent stage, returns to thinking mode, and resumes processing reasoning segments.
Figure~\ref{fig:architecture} shows the state diagram of \model.

In summary, \model functions as an incremental, asynchronous, speech-oriented summarizer of the reasoning output. Since \model performs no reasoning itself, it can be implemented with a more compact model compared to the reasoning LLM.

\captionsetup[algorithm]{font=small}  
\begin{algorithm}[!t]
    \small
    \caption{\framework with \model}
    \label{alg:revert}
    \begin{algorithmic}[1]
        \Require a trained \model $p_\theta$, a reasoning model $q$, user query tokens $\gQ$, a set of delimiters $\gD$.
        \vspace{0.5em}
        \Function{\colorbox{red!10}{Think}}{$q$, $\gQ$}
            \State initialize $i \leftarrow 0$ 
            \Repeat 
                \State generate $r_{i} \sim q(\cdot\mid \gQ, r_{<i})$
                \State send $r_{i}$ to the verbalizer
                \State $i \leftarrow i + 1$
            \Until {$r_{i-1} = \meos$}
        \EndFunction

        \vspace{0.5em}
        \Function{\colorbox{yellow!30}{Verbalize}}{$p_{\theta}, \gQ$}
            \State set the current state $\gS$ as \textit{thinking mode} $\gS_{T}$
            \State initialize a context $\gC \leftarrow \gQ$
            \While {reasoning is not complete}
                \State receive texts from the reasoning model.
                \State process these texts into segment $\gR$ with $\gD$.
                \If {$\gS$ is in \textit{thinking mode} ($\gS_{T}$)}
                    \State update $\gC \leftarrow (\gC, \gR)$.
                    \State sample $s \sim p_\theta(\cdot\mid\gC)$ \Comment{\con or \bov}
                    \If {$s = \mbov$}
                        \State transition state $\gS$ to \textit{verbalizing mode}.
                    \EndIf
                \EndIf
                \If {$\gS$ is in \textit{verbalizing mode} ($\gS_{V}$)}
                    \State update $\gC \leftarrow (\gC, \mbov)$ \Comment{Begin verbalization}
                    \State initialize the verbalization buffer $\gV \leftarrow ()$.
                    \Repeat
                        \State generate $v \sim p_{\theta}(\cdot\mid \gC)$.
                        \State update context: $\gC \leftarrow (\gC, v)$.
                        \State $\gV \leftarrow (\gV, v)$
                    \Until {$v = \meov$} \Comment{End of verbalization}
                    \State transition state $\gS$ to \textit{thinking mode}.
                    \State send $\gV$ to the TTS model.
                \EndIf
            \EndWhile
        \EndFunction
    \end{algorithmic}
\end{algorithm}
\subsection{\textsc{Speak}}

In the \Speak stage, we convert the utterances to speech using a TTS model. Specifically, we employ a TTS model that supports both streaming input and output, allowing the system to process streaming outputs from the \Verbalize stage and play the generated speech with minimal delay for the user.

\begin{figure*}[tb!]
\centering
\includegraphics[width=\linewidth]{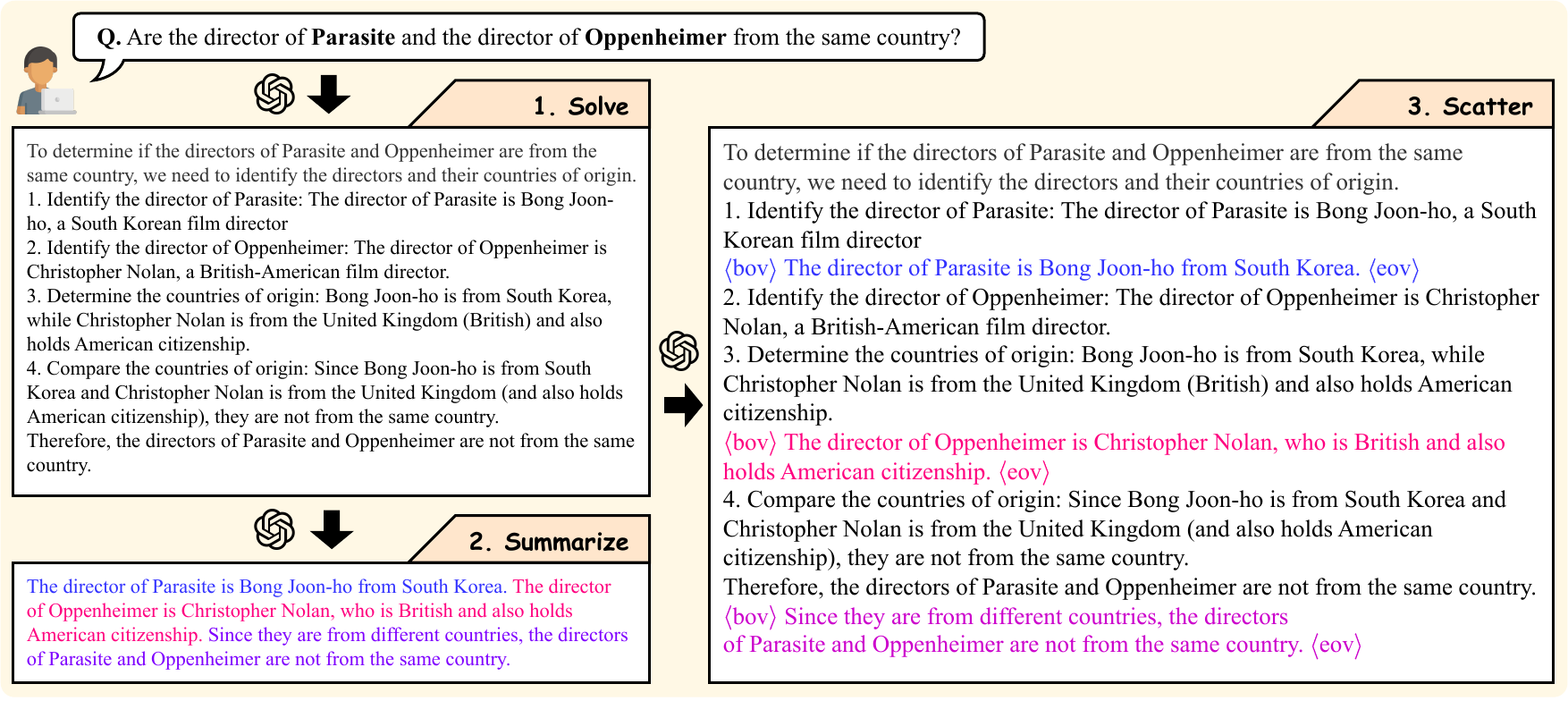}
\caption{Data construction pipeline for training \model. Given a question, the process involves three steps: (1) \textbf{Solve}: Generate a step-by-step reasoning process to derive the answer, (2) \textbf{Summarize}: Extract key components of the reasoning and rewrite them as speech-friendly utterances. (3) \textbf{Scatter}: Insert each utterance immediately after the corresponding reasoning segment, creating an interleaved sequence of internal reasoning and verbal explanations.}
\label{fig:data_construction}
\vspace{-0.3cm}
\end{figure*}

\subsection{\model Training}

Since the reasoning LLM and streaming TTS models remain frozen, we describe only the training procedure for the \model model. We below discuss the training data format, the dataset construction pipeline, and the training objective.

\paragraph{Training Data}
Each training example comprises a user query $\gQ$ and the corresponding response $\gX$. Since \model performs incremental summarization of reasoning steps, the training data must be structured such that summaries are interleaved with their respective reasoning segments. Formally, $\gX$ is represented as
\begin{align}
    \gX &= \big[\gX_{1}, \dots \gX_{n}\big], \\
    \gX_{k} &= \big[\gR_{k} \ \mbov \ \gV_{k} \ \meov\big],
\end{align}
where $\gR_k$ is the segments of the $k$-th reasoning step, and $\gV_k$ is the verbalized text, enclosed by \bov and \eov tokens, as a speech-friendly summary of $\gR_k$.
Sometimes, $\gR_k$ consists of multiple reasoning segments, denoted as $\gR_k = [\gR_{k}^{1}, \dots, \gR_{k}^{m_k}]$, where each segment is separated by delimiters $\gD$ (\ie, newline), and $m_k$ indicates the total number of segments.

\paragraph{Dataset Construction Pipeline}
Because no publicly available datasets conform to the required format, we propose a simple LLM-based pipeline to generate a dataset in our desired format with a standard QA dataset as input. Figure~\ref{fig:data_construction} presents an overview of the proposed pipeline. The pipeline consists of three steps: \textit{solve}, \textit{summarize}, and \textit{scatter}. In the solve step, the reasoning LLM solves the user query using a standard chain-of-thought reasoning. In the summarize step, we generate a speech-friendly summary for the generated reasoning output. In the scatter step, we scatter the summaries across the reasoning process such that each summary appears immediately after its associated reasoning step, enclosed by $\mbov$ and $\meov$ tokens. We use the output of the scatter step as the training data for \model.
For all three steps, we employ \gptf as the processing model. More detailed procedures and prompts are provided in Appendix \ref{sec:app_dataset}.

\paragraph{Objective}
The training procedure for \model closely follows standard LLM finetuning. We begin by initializing \model with a pretrained LLM and finetune it using cross-entropy loss on the next-token prediction task, applied selectively to the training data described above. Importantly, since \model is not required to perform the reasoning process itself, we compute the loss only within the verbalization segments of each sequence. For positions outside these verbalization segments, the model is trained to predict a special \con token, signaling that it is still in the thinking mode.

Formally, let $\mathcal{I}_{\textsc{Verbal}}$ denote the set of token positions within verbalization segments, the set of token positions spanning from \bov{} to \eov{}, inclusive. Conversely, let $\mathcal{I}_{\Think}$ represent the set of token positions outside $\mathcal{I}_{\textsc{Verbal}}$, corresponding to the tokens used for LLM reasoning. Then, the total loss is
\begin{align}
\mathcal{L}(\theta)  = & - \sum_{i \in \mathcal{I}_{\textsc{Verbal}}} {\log p_{\theta}(x_i \mid \gQ, x_{<i})} \\
& - \sum_{i \in \mathcal{I}_{\Think}} \log p_{\theta}(\mcon \mid \gQ, x_{<i}).
\end{align}
Here, $x_i$ is the $i$-th token in the response sequence $\gX$, and $p_\theta$ is the model's output probability.

\section{Experiments} \label{sec:experiments}
We evaluate the effectiveness of our \framework framework and the verbalizer model across multiple experimental settings.
Full details are provided in Appendix~\ref{sec:app_training}.

\subsection{Models}

We evaluate two versions of \framework: (1) \textbf{TVS (\SEQ)}, which performs reasoning followed by verbalization sequentially; and (2) \textbf{TVS (\model)}, in which the \model model incrementally verbalizes the reasoning outputs.

For comparison, we include several baselines based on the \textsc{Think-Speak} framework:
(1) Chain-of-Thought (\textbf{CoT}) employs a standard zero-shot chain-of-thought prompting technique to elicit step-by-step reasoning;
(2) Speech-Friendly Prompting (\textbf{SFP}) applies prompting strategies to encourage the model to generate concise, speech-appropriate outputs, following the guidelines established by \citet{speechworthy}; and
(3) Speech-Friendly Finetuning (\textbf{SFF}) uses a finetuned model to directly produce speech-friendly responses. For finetuning, we use the same dataset as our model, but replace the output of the scatter step with that of the summarize step.
Additionally, we include Qwen2.5-Omni-7B~\citep{qwen2omni}, an end-to-end spoken dialogue system finetuned to produce speech-friendly outputs, as a baseline for comparative analysis.

For the \Think model, we experiment with multiple LLMs, specifically \qwen~\citep{qwen2}, \llama~\citep{llama}, and \gpt~\citep{gpt4o}. For \SEQ and \model model, we use \qwenv~\citep{qwen2} as the base model and fine-tune it. For all models except \qweno, we employ gpt-4o-mini-tts~\citep{gpttts} as the speak model to convert textual responses into speech.

\subsection{Datasets}

We consider the following three datasets for our evaluation setup:
(1) \gsm8k~\citep{gsm8k} focuses on arithmetic reasoning, based on grade-school level math problems. The solutions are generally straightforward and linear, involving simple, easy-to-follow steps without complex mathematical elements;
(2) \2wiki~\citep{2wikimultihop} requires multi-hop retrieval of Wikipedia documents to answer a question. While the dataset is not primarily designed to assess complex reasoning, multi-hop QA in a closed-book setting demands step-by-step reasoning abilities; and
(3) \scibench~\citep{scibench} assesses college-level scientific problem-solving abilities. The solutions are often involve complex equations, formulas, and other components that are not easily communicated verbally.

We construct the training set of \model as a subset of examples from the \gsm8k and \2wiki training sets.
\scibench remains unseen during training and serves to evaluate the model’s out-of-domain generalization capability.

\subsection{Evaluation Procedure and Measures}

\paragraph{Automatic Reasoning Evaluation} We evaluate the reasoning capabilities of dialogue systems. Each system generates responses to the provided questions, and we assess the correctness of the final outputs using an LLM-as-a-judge framework.
We report the accuracy for this evaluation.

\paragraph{Automatic Speech-Friendliness Evaluation}
We evaluate whether the responses from each system are suitable for verbal delivery. We adopt the four metrics also used by \citet{speechworthy}: (1) Word count (WC) measures the overall conciseness of the response and is computed using simple whitespace delimitation; (2) Flesch Reading Ease (FRE) score assesses text readability based on the number of syllables per word and words per sentence. Although not directly related to speech, the FRE score is correlated with listenability; (3) Dependency depth (DD) is the maximum depth of the response dependency tree computed by Spacy dependency parser\footnote{\url{https://spacy.io/api/dependencyparser}}. DD helps assess the sentence complexity; (4) Nonvocalizable character count (NV) evaluates the appropriateness of the response for verbal delivery by identifying the presence of nonvocalizable content.

\begin{table}[t!]
    \centering
    \small
    \setlength\tabcolsep{7pt}
    \begin{tabularx}{\linewidth}{X}
    \toprule
       \textbf{Criteria} \\
    \midrule
       \textbf{Naturalness:} Whether the response sounds like something a human would naturally say in conversation. \\
       \cmidrule(lr){1-1}
       \textbf{Conciseness:} Whether the response delivers essential information without unnecessary verbosity. \\
       \cmidrule(lr){1-1}
       \textbf{Understandability:} How easily the response can be comprehended when spoken aloud. \\
       \cmidrule(lr){1-1}
       \textbf{Overall Quality:} Overall impression of the response's quality and suitability for spoken delivery. \\
    \bottomrule
    \end{tabularx}
    \caption{Criteria for human evaluation of spoken responses.
    Detailed descriptions are available in Appendix~\ref{sec:app_human_evaluation}.}
    \label{fig:eval_criteria}
    \vspace{-0.3cm}
\end{table}

\paragraph{\model Latency Evaluation}
We measure the time-to-response of the \framework framework and evaluate the effectiveness of \model in latency reduction. Since we use a streaming TTS model, we focus on the time required to generate the first spoken output, specifically $\mathbf{T}_{1}$, the time taken for the system to enter the verbalizing mode after receiving the user’s query, and $\mathbf{T}_{2}$, the additional time required to produce the first verbalized segment after verbalization has started. We report latencies at the 50th percentile with Qwen2.5-3B-Instruct as the verbalizer. All experiments are conducted on the GSM8K dataset using the PyTorch \texttt{transformers} library with bfloat16 precision on an NVIDIA A6000 GPU.

\begin{table*}[tb!]
    \centering
    \scriptsize
    \setlength\tabcolsep{5pt}
    \begin{tabularx}{0.95\textwidth}{lYYYYYYYYY}
    \toprule
        & \multicolumn{3}{c}{(a) Accuracy (\%)} & \multicolumn{4}{c}{(b) Speech-suitability} & \multicolumn{2}{c}{(c) Latency (s)} \\
        \cmidrule(lr){2-4} \cmidrule(lr){5-8} \cmidrule(lr){9-10}
        Models & \gsm8k & \mhqa & \scibench & WC(↓) & FRE(↑) & DD(↓) & NV(↓) & $\mathrm{T}_1$ & $\mathrm{T}_2$ \\
    \midrule
        \textbf{\textit{\qweno}} & 84.53 & 14.30 & 20.95 & 101.7 & 90.90 & 5.24 & 0.78 & - & - \\
    \midrule
        \multicolumn{10}{l}{\textit{\textbf{Qwen2.5-7B-Instruct}}} \\
        \cmidrule(lr){1-10}
        ~~Chain-of-Thought          & 92.72 & 30.00 & 50.72 & 153.5 & 69.22 & 6.23 & 67.11 & 0.0 & 0.64  \\
        ~~Speech-Friendly Prompting & 87.57 & 26.60 & 45.09 & 87.11 & 84.97 & 5.45 & 11.04 & 0.0 & 0.46  \\
        ~~Speech-Friendly Finetuning & 68.69 & 32.70 & 21.97 & 44.90 & 88.32 & 4.28 & 0.035 & 0.0 & 0.47  \\
        \cmidrule(lr){1-10}	
        ~~\textbf{TVS (\SEQ)}       & 93.18 & 29.75 & 47.40 & 42.15 & 88.71 & 4.23 & 0.005 & 8.08 & 0.43 \\
        ~~\textbf{TVS (\model)}     & 92.65 & 30.00 & 47.25 & 44.02 & 88.40 & 4.21 & 0.024 & 2.72 & 0.45	\\
    \midrule
        \multicolumn{10}{l}{\textit{\textbf{Llama-3.1-8B-Instruct}}} \\
        \cmidrule(lr){1-10}
        ~~Chain-of-Thought           & 85.44 & 17.95 & 28.32 & 194.5 & 69.86 & 6.42 & 9.079 & 0.0 & 0.77 \\
        ~~Speech-Friendly Prompting & 83.70 & 16.45 & 22.83 & 101.7 & 87.99 & 5.33 & 2.887 & 0.0 & 0.34 \\
        ~~Speech-Friendly Finetuning & 65.13 & 42.50 & 14.02 & 48.05 & 88.43 & 4.25 & 0.034 & 0.0 & 0.35 \\
        \cmidrule(lr){1-10}
        ~~\textbf{TVS (\SEQ)} & 85.44 & 22.25 & 26.01 & 43.94 & 88.72 & 4.27 & 0.026 & 7.19 & 0.44 \\        
        ~~\textbf{TVS (\model)} & 85.29 & 19.10 & 27.80 & 44.95 & 88.89 & 4.20 & 0.043 & 2.67 & 0.44 \\
    \midrule
        \multicolumn{10}{l}{\textit{\textbf{gpt-4o-mini-2024-07-18}}} \\
        \cmidrule(lr){1-10}
        ~~Chain-of-Thought & 94.84 & 39.60 & 55.64 & 175.4 & 67.40 & 6.37 & 74.69 & - & - \\
        ~~Speech-Friendly Prompting & 87.26 & 34.40 & 34.54 & 73.09 & 82.45 & 5.14 & 0.215 & - & - \\
        \cmidrule(lr){1-10}
        ~~\textbf{TVS (\SEQ)} & 94.77 & 39.75 & 53.26 & 43.83 & 88.48 & 4.27 & 0.008 & - & - \\
        ~~\textbf{TVS (\model)} & 94.69 & 39.55 & 53.32 & 45.92 & 88.39 & 4.25 & 0.019 & - & - \\
    \bottomrule
    \end{tabularx}
    \caption{Main results comparing different baselines and our proposed methods (TVS (\SEQ) and (\model) ) across three base \Think models. We report (a) task accuracy on \gsm8k, \2wiki, and \scibench; (b) speech-suitability scores using word count (WC), Flesch Reading Ease (FRE), dependency depth (DD), and number of non-vocal characters (NV); and (c) generation latency ($\mathrm{T}_1$, $\mathrm{T}_2$) at the \nth{50} percentile. Speech-suitability scores and latencies are computed on the \gsm8k test set. By decoupling thinking and verbalizing (TVS), we substantially improve speech-friendliness while preserving reasoning capabilities of the Chain-of-Thought baseline. Furthermore, the use of the \model{} model significantly reduces latency. Results of speech-suitability evaluation on additional datasets are presented in Appendix~\ref{sec:app_analysis}.}
    \label{tab:main_results}
    \vspace{-0.3cm}
\end{table*}

\paragraph{Human Evaluation}
We conduct a human evaluation where Amazon Mechanical Turk annotators rate system responses using a 5-point Likert scale across four criteria: naturalness, conciseness, understandability, and overall quality. Table \ref{fig:eval_criteria} provides the definitions of each criterion. We randomly sample 60 examples, 20 from each dataset, and collect annotations from three independent raters per example. Unlike previous evaluations that rely on textual assessment, this evaluation is \textit{speech-based}.

\section{Results and Discussion} \label{sec:results}
\subsection{Does speech-friendliness compromise models' reasoning capabilities?}
Table~\ref{tab:main_results} presents the results of the automatic evaluations for the \framework model and the baseline systems. In most cases, the chain-of-thought strategy achieves the highest reasoning benchmark accuracies within each \Think model category, but demonstrates the lowest performance in speech-suitability evaluations.
This indicates that the chain-of-thought strategy exhibits highly polarized performance with respect to reasoning capabilities and speech-friendliness.

Therefore, we apply two most widely used solutions to these issues: prompting and finetuning. While the speech-friendly prompting yields only a minimal decrease in reasoning accuracies, it ignores the instructions when faced with challenging questions and resorts to chain-of-thought reasoning, thereby harming its speech-suitability scores. An example in Table~\ref{tab:app_qual_2wiki} and human evaluation in Table~\ref{tab:human_eval_full} reveal similar issues. Despite receiving the highest overall scores on \2wiki, its scores, especially the conciseness score, drop significantly on \gsm8k and \scibench. \qweno also exhibits a similar trend, where its speech-friendliness diminishes with rising problem difficulty.

In contrast, the speech-friendly finetuning system receives high speech-friendliness scores but low reasoning benchmark scores. In other words, it yields highly intelligible responses but not intelligent ones. Notably, the system achieves the highest scores on the \2wiki dataset. We attribute this to the model acquiring additional knowledge during training, as the dataset does not strictly separate train set and development set knowledge bases. Therefore, the high score is likely unrelated to the system’s reasoning capabilities.

These findings highlight a fundamental trade-off within the two-stage paradigm: optimizing for reasoning capability tends to degrade speech-suitability, and vice versa.

\begin{table*}[t!]
    \centering
    \scriptsize
    \setlength\tabcolsep{2pt}
    \begin{tabularx}{\textwidth}{lYYYYYYYYYYYY}
    \toprule
        & \multicolumn{4}{c}{\gsm8k} & \multicolumn{4}{c}{\2wiki} & \multicolumn{4}{c}{\scibench}\\
        \cmidrule(lr){2-5} \cmidrule(lr){6-9} \cmidrule(lr){10-13}
\textbf{Method}  & Natu. & Conc. & Unde. & Over. & Natu. & Conc. & Unde. & Over. & Natu. & Conc. & Unde. & Over. \\
    \midrule
CoT          & \mstd{4.55}{0.09} & \mstd{3.72}{0.11} & \mstd{4.48}{0.09} & \mstd{4.28}{0.09}
             & \umstd{3.92}{0.16} & \umstd{2.92}{0.18} & \umstd{4.40}{0.11} & \umstd{3.53}{0.15}
             & \mstd{4.25}{0.12} & \mstd{3.10}{0.16} & \mstd{3.75}{0.12} & \mstd{4.08}{0.10} \\
SFP          & \mstd{4.32}{0.12} & \mstd{3.67}{0.15} & \mstd{4.52}{0.12} & \mstd{4.17}{0.12}
             & \bmstd{4.47}{0.10} & \bmstd{4.23}{0.12} & \bmstd{4.65}{0.08} & \bmstd{4.35}{0.10}
             & \mstd{4.20}{0.11} & \umstd{3.02}{0.18} & \bmstd{3.90}{0.11} & \mstd{4.03}{0.09} \\
SFF          & \bmstd{4.62}{0.09} & \mstd{4.33}{0.11} & \mstd{4.52}{0.10} & \mstd{4.33}{0.10}
             & \bmstd{4.47}{0.12} & \mstd{4.17}{0.13} & \mstd{4.60}{0.10} & \mstd{4.32}{0.12}
             & \mstd{4.23}{0.15} & \mstd{4.00}{0.12} & \mstd{3.80}{0.12} & \mstd{4.10}{0.10} \\
\qweno       & \umstd{3.40}{0.14} & \umstd{3.50}{0.14} & \umstd{3.72}{0.15} & \umstd{3.50}{0.14}
             & \mstd{4.03}{0.14} & \mstd{4.18}{0.10} & \mstd{4.47}{0.10} & \mstd{4.10}{0.11}
             & \umstd{4.00}{0.16} & \mstd{3.48}{0.15} & \umstd{3.70}{0.14} & \umstd{3.93}{0.14} \\
    \midrule
\textbf{TVS (\SEQ)}  & \bmstd{4.62}{0.09} & \mstd{4.40}{0.10} & \bmstd{4.55}{0.09} & \bmstd{4.37}{0.09}
                   & \mstd{4.35}{0.12} & \mstd{4.10}{0.11} & \mstd{4.58}{0.10} & \mstd{4.33}{0.10}
                   & \mstd{4.40}{0.10} & \mstd{4.02}{0.11} & \mstd{3.82}{0.13} & \bmstd{4.28}{0.09} \\
\textbf{TVS (\model)}& \mstd{4.58}{0.10} & \bmstd{4.45}{0.11} & \mstd{4.23}{0.13} & \mstd{4.23}{0.12}
                   & \mstd{4.39}{0.13} & \mstd{4.21}{0.13} & \mstd{4.63}{0.10} & \mstd{4.18}{0.11}
                   & \bmstd{4.45}{0.09} & \bmstd{4.07}{0.12} & \mstd{3.87}{0.14} & \bmstd{4.28}{0.09} \\
    \bottomrule
    \end{tabularx}
    \caption{
        Human evaluation scores for spoken responses using a 5-point Likert scale. "Natu.", "Conc.", "Unde.", and "Over." denote Naturalness, Conciseness, Understandability, and Overall Quality, respectively. Each cell contains the mean and standard error of the ratings across three datasets. \textbf{Bold} indicates the highest score in each column, and \underline{underline} indicates the lowest.
    }
    \vspace{-0.25cm}
    \label{tab:human_eval_full}
\end{table*}

\subsection{How does the explicit verbalization stage affect performance?}
\label{sec:tvs_effectiveness}

While the \framework framework, by design, should mirror the accuracy scores of the \Think model’s chain-of-thought strategy, we observe a slight decrease in accuracy on the \scibench dataset. We attribute this to two possible factors: (1) out-of-domain characteristics and (2) inherent task difficulty. However, even with the drop in accuracy, both versions of our framework vastly outperform other baselines.

We also observe an anomalous result on the \2wiki dataset for the \llama \Think model, where both the \SEQ and \model variants outperform the chain-of-thought strategy. We attribute this to the same factor identified in the speech-friendly finetuning strategy issue, as all three systems share the same target text in the training data.

For speech-suitability measures, both \SEQ and \model outperform all other baselines in automatic evaluation. In human evaluation, we analyze the results for each dataset. On \2wiki, all systems achieve high scores. In \gsm8k and \scibench{}, all systems show a performance drop in conciseness and understandability. Regardless, \SEQ and \model consistently rank as the top two models in terms of naturalness, conciseness, and overall quality criteria.

In summary, the introduction of the \Verbalize stage in our framework enables exceptional speech-friendliness with minimal compromise in the reasoning capabilities of the \Think model.

\subsection{When should I use \model over \SEQ?}

As stated in Section~\ref{sec:tvs_effectiveness}, both \SEQ and \model perform well across different datasets, with minimal differences in their effectiveness as verbalizers. The primary distinction between the two models lies in their latency. Specifically, \SEQ waits for the reasoning process to complete before verbalizations, which requires approximately 8.08 seconds, as shown in Table~\ref{tab:main_results}(c). Such latency is unsuitable for real-time spoken conversation settings.

Conversely, \model incrementally processes verbalizable segments before the reasoning process is complete, receiving the first segment in an average of 2.72 seconds, a 66\% reduction in latency compared to \SEQ. In voice-interface conversations, this latency can be effectively masked by brief filler phrases such as “Let me think,” making it acceptable for real-time applications.

Therefore, \model achieves performance comparable to \SEQ while significantly reducing latency, suggesting that \model is preferable for most real-time applications.
\begin{table}[t!]
    \centering
    \small
    \setlength\tabcolsep{3pt}
    \begin{tabularx}{\linewidth}{lccYY}
    \toprule
        \multirow{2}{*}{\model{} size}
        & \multicolumn{2}{c}{Accuracy (\%)}
        & \multicolumn{2}{c}{Speech-suitability} \\
    \cmidrule(lr){2-3} \cmidrule(lr){4-5}
        & GSM8K & SciBench & WC (↓) & FRE (↑) \\
    \midrule
        7B   & 92.7 & 50.7 & –    & –    \\
    \midrule
        3B   & 92.7 & 47.3 & 44.0 & 88.4 \\
        1.5B & 92.7 & 46.8 & 45.3 & 88.9 \\
        0.5B & 91.4 & 42.1 & 44.2 & 88.6 \\
    \bottomrule
    \end{tabularx}
    \vspace{-0.2cm}
    \caption{Comparison of verbalization abilities across different \model{} model sizes. Speech-suitability scores, consisting of word count (WC) and Flesch Reading Ease (FRE), are calculated on GSM8K.}
    \label{tab:res_ablation}
    \vspace{-0.3cm}
\end{table}

\subsection{Does size matter?}

We discuss the effect the \model model size has on its performance. Table~\ref{tab:res_ablation} illustrates the performance of three differently sized \model models: Qwen2.5-0.5B-Instruct, Qwen2.5-1.5B-Instruct, and Qwen2.5-3B-Instruct. The results indicate that the performance loss from decreasing model size is more pronounced on SciBench than on GSM8K, likely reflecting the greater task difficulty of the former dataset. Notably, the speech-suitability scores remain stable despite reductions in model size.

In conclusion, although model size affects \model's performance, the degradation is not substantial. This suggests that smaller \model models remain a viable option in low-resource settings.

\section{Conclusion}
In this work, we address a critical gap between text readability and speech-suitability in spoken dialogue system responses. We present the \framework{} framework, which decouples reasoning from verbalization to achieve both reasoning accuracy and speech-friendliness. Extensive automatic and human evaluations show that our framework enhances speech-suitability with minimal compromise of reasoning capability across benchmarks. Additionally, we introduce the \model{} model for incremental verbalization, which reduces latency compared to the sequential approach. Extending the framework to multi-turn or full-duplex interactions presents a promising avenue for future research.

\section{Limitations}
While our framework shows promising results, it has several limitations. First, it focuses on single-turn conversational settings and does not support multi-turn or full-duplex interactions, where reasoning and verbalization may occur in parallel with multiple user interactions. Extending the framework to handle such interactive scenarios remains an important direction for future work. Second, the current verbalization model does not allow control over the level of explanation detail. Adding support for adjustable granularity, ranging from brief summaries to step-by-step explanations, could improve adaptability to different user needs. Third, our work focuses on chain-of-thought reasoning, but extending it to other test-time computation methods with intermediate traces, such as multi-step retrieval or tool use, could broaden its applicability.

\section{Potential Risks}
Our framework introduces no additional epistemic or safety risks beyond those already present in the underlying reasoning model. This is because the verbalization model is designed solely to rephrase the outputs of a frozen, pretrained reasoning LLM into speech-friendly language without altering their content or logic. It performs no independent reasoning, decision-making, or content generation beyond linguistic reformulation. Consequently, factual inaccuracies, biases, or harmful outputs originate entirely from the reasoning model. The verbalization stage merely translates those outputs into a form more suitable for spoken communication. Thus, the overall risk profile of the system is bounded by that of the underlying reasoning model, and our model introduces no novel vulnerabilities.

\section*{Acknowledgements}
We thank the reviewers for the valuable feedback. This work was supported by the Institute of Information \& Communications Technology Planning \& Evaluation (IITP) grant funded by the Korea government (MSIT) under the following projects: No. RS-2025-25442338 (AI Star Fellowship Support Program), No. RS-2022-II220156 (Fundamental research on continual meta-learning for quality enhancement of casual videos and their 3D metaverse transformation), No. RS-2025-02263841 (Development of a Real-time Multimodal Framework for Comprehensive Deepfake Detection Incorporating Common Sense Error Analysis), and IITP-2025-RS-2024-00437633 (IITP–ITRC). It was also supported by R\&BD Program (CD200024) through the Seoul Business Agency(SBA)
funded by The Seoul Metropolitan Government. This research was also conducted as part of the Sovereign AI Foundation Model Project(Data Track), organized by the Ministry of Science and ICT(MSIT) and supported by the National Information Society Agency(NIA), S.Korea (2025-AI Data-wi43). Gunhee Kim is the corresponding author.

\section{Ethical Statement}
All models, datasets, and other artifacts used in this work are released under licenses that permit research use. Our usage of these resources is consistent with both the terms of their licenses and the intended purposes specified by their creators.

\bibliography{main}

\clearpage
\appendix
\section{Dataset} \label{sec:app_dataset}
This section provides a detailed procedure for generating our training dataset. First, we sample a set of raw question-answer pairs from the \gsm8k and \2wiki.

From the \gsm8k training set, we use all 7,473 examples. From the \2wiki dataset, we sample 1,000 examples from each of the 4 data types: inference, comparison, bridge\_comparison, and compositional.

\subsection{Solve, Summarize, Scatter}

\paragraph{Solve} In this step, we simply induce step-by-step reasoning process, using standard zero-shot chain-of-thought prompting.

\paragraph{Summarize} In this step, we generate a summary of the reasoning process from \textit{solve}. We impose the following constraints on the resulting summary:
\begin{itemize}
	\item The summary must contain all essential information from the reasoning process.
	\item The summary must follow the same logical progression as the reasoning process.
	\item The summary must not repeat information provided in the question.
	\item The summary must be speech-friendly and free of complex sentences or hard-to-read words.
\end{itemize}

Because enforcing all constraints simultaneously in a single instruction yields suboptimal results, we adopt a progressive approach, providing the language model with one constraint at a time.

\paragraph{Scatter} In this step, we distribute the summary throughout the reasoning process, placing each summary segment immediately after its corresponding reasoning segment. To encourage fine-grained control over the placement of summary segments, we manually label 16 samples and use them as few-shot examples.

\begin{table*}[t!]
    \centering
    \scriptsize
    \begin{tabularx}{\textwidth}{l|X}
    \toprule
    \multicolumn{2}{l}{\textbf{Prompts for Data Construction}} \\
    \midrule
    \textbf{Solve} & Provide a step-by-step reasoning process before arriving at the final answer. \\
    \midrule
    \textbf{Summarize} & Write a clear, concise, and speech-friendly summary of the provided analysis. Ensure the summary flows naturally when read aloud, avoiding complex sentences, mathematical equations or awkward phrasing. Follow the analysis's logical progression, presenting key points in the same order and context. Include only essential steps, omitting unnecessary details, boilerplate text, or repetitions. The general tone of the summary must match that of the original analysis. The summary must be appropriate as the response to the original question. Do not apply LaTeX or markdown formatting in your summary.\newline\newline
    First, extract the essential information from the analysis. \\
    \cmidrule(rr){2-2}
& Now, make sure the logical progression of the summary matches the order of the analysis, without adding or removing information. \\
\cmidrule(rr){2-2}
& Now, remove all parts that are repeating the information from the original question. \\
\cmidrule(rr){2-2}
& Now, make sure the content is speech-friendly be breaking up complex sentences and avoiding long and hard-to-read words. \\
\midrule
\textbf{Scatter} & Combine an analysis with its corresponding summary by interleaving summary segments with relevant analysis portions.\newline
\newline
Ensure the summary segments immediately follow the equivalent content within the analysis and are enclosed with `<bov>` and `<eov>` tags.\newline
\newline
\# Steps\newline
\newline
1. **Receive analysis and summary**: Identify sections in both the analysis and the summary.\newline
2. **Match content segments**: Pair each summary segment with the corresponding analysis segment to ensure logical flow.\newline
3. **Interleave content**: Insert summary segments after the matching sections of the analysis, surrounded by `<bov>` and `<eov>` tags.\newline
\newline
\# Output Format\newline
\newline
The output should maintain the structure of the analysis, with summary segments appropriately interleaved. Each summary segment must be enclosed in `<bov>` and `<eov>` tags immediately after the equivalent analysis content.\newline
\newline
\# Notes\newline
\newline
- Maintain the logical sequence of both the analysis and summary.\newline
- Ensure clarity in how summary elements align with the analysis.\newline
- Avoid rephrasing the given segments; use them verbatim for consistency and accuracy.\newline
- The summary and analysis must not change from the original text. \\
    \bottomrule
    \end{tabularx}
    \caption{Prompts for dataset construction.}
    \label{tab:data-prompts}
\end{table*}

\section{Experimental Setup} \label{sec:app_setup}
\subsection{Prompts} \label{sec:app_prompts}
This section outlines the specific prompts used in our experiments, including those for baseline methods and our proposed verbalizer.
For the CoT reasoning experiments, we adopt the system prompt illustrated in Figure~\ref{fig:cot-prompt}. In the case of speech-friendly prompting and finetuning, we follow the instruction template shown in Figure~\ref{fig:sfp-prompt}.
Our proposed verbalizer (both the \SEQ and \model) uses the prompt presented in Figure~\ref{fig:revert-prompt}. For experiments involving Qwen2.5-Omni-7B, we employ the default system prompt provided by the model.

\begin{figure}[h!]
    \centering
    \begin{tcolorbox}
    [colback=gray!5!white,colframe=black!75!black,title={Chain-of-Thought Prompting (CoT)},fontupper=\ttfamily\small]
        You are a helpful assistant that provides a step-by-step reasoning process before arriving at the final answer.
    \end{tcolorbox}
    \vspace{-0.3cm}
    \caption{A system prompt designed for chain-of-thought (CoT) prompting.}
    \label{fig:cot-prompt}
\end{figure}

\begin{figure}[h!]
    \centering
    \begin{tcolorbox}
    [colback=gray!5!white,colframe=black!75!black,title={Speech-Friendly Prompting (SFP)},fontupper=\ttfamily\small]
        You are a voice assistant that responds in a way that is easy to understand when spoken aloud.
        Your responses should be concise, clear, and listener-friendly.
        Avoid using equations, LaTeX, or complex symbols that are hard to pronounce or understand in speech.
    \end{tcolorbox}
    \vspace{-0.3cm}
    \caption{A system prompt designed for speech-friendly prompting (SFP) or finetuning (SFF).}
    \label{fig:sfp-prompt}
\end{figure}

\begin{figure}[h!]
    \centering
    \begin{tcolorbox}
    [colback=gray!5!white,colframe=black!50!black,title={\model},fontupper=\ttfamily\small]
        Your task is to provide step-by-step reasoning with incremental,     speech-friendly summaries. You will be given a question and must reason through it step by step.
        Each time you generate the token <bov>, provide a clear and concise summary of the reasoning so far, suitable for spoken delivery.
        Do not include equations, LaTeX, or complex symbols in these summaries.
        Conclude each summary with the token <eov>.
        Ensure each summary connects naturally to the one before it.
    \end{tcolorbox}
    \vspace{-0.3cm}
    \caption{A system prompt designed for our verbalizer.}
    \label{fig:revert-prompt}
\end{figure}

\subsection{Training} \label{sec:app_training}
We finetune our verbalization model for \SEQ and \model model using Qwen2.5-3B-Instruct with full-parameter optimization. All models are trained for one epoch with 4 A6000GPU, totaling 1.3k steps (within 1 hour) with a batch size of 8. For optimization, we employ the AdamW optimizer with a learning rate of $2\times 10^{-5}$, a cosine learning rate scheduler, and a warmup ratio of 0.1. The optimizer parameters are set to $\beta_1 = 0.9$, $\beta_2 = 0.999$, with a weight decay of 0.1.
For speech-friendly finetuning (SFF), we finetune Qwen2.5-7B-Instruct and Llama-3.1-8B-Instruct using LoRA with $r=16$ and $\alpha=16$. and other training configurations are kept identical to those described above.

\subsection{Inference} \label{sec:app_inference}
We use top-$p$ sampling with a temperature of 0.1 and a nucleus probability $p=0.95$ for all response generation for inference. For the \model model, we employ greedy decoding for next-token prediction to determine whether to initiate verbalization (i.e., generate the \bov{} or \con{} token). Upon receiving the final reasoning segment, the verbalizer is manually appended with the \bov{} token rather than relying on sampling.

\subsection{LLM-Based Answer Verification} \label{sec:app_answer_verification}
We utilize an LLM-based answer verification method to overcome the limitations of rule-based evaluation. In the context of speech-friendliness, responses should be clear, natural, and easily understandable, which means they may not always conform to a specific format or template. Such characteristics render exact matching and rule-based answer extraction unreliable.

Therefore, we use \gptf to automatically assess answer correctness. As illustrated in Figure~\ref{fig:answer-verification-prompt}, the verifier is prompted with the question, a model-generated response, and the corresponding ground-truth answer.

\begin{figure*}[t!]
    \centering
    \begin{tcolorbox}
    [colback=gray!5!white,colframe=black!75!black, fontupper=\ttfamily\small, fontlower=\ttfamily\small, title=Answer Verification Prompt]
    \textbf{[SYSTEM]} \\
    Given a ground-truth answer and a submitted answer, tell me if the submitted answer is correct. Disregard any units, formatting, etc. In your explanation, first extract the final answer of the submitted answer and then compare it with the ground truth. \\
    \\
    You must only use the ground truth answer to determine the correctness of the submitted answer. The validity of the ground truth answer must not be questioned. Note that the submitted answer must directly and explicitly answer the question. Any implicit answering should be considered incorrect.
    \tcblower{
    \textbf{[USER]} \\
    Question:\\
    \textcolor{blue}{\{question\}}\\
    \\
    Ground-Truth Answer: \\
    \textcolor{blue}{\{ground\_truth\}}\\
    \\
    Submitted Answer:\\
    \textcolor{blue}{\{submitted\_answer\}}
    }
    \end{tcolorbox}
    \vspace{-0.3cm}
    \caption{The prompt for LLM-based answer verification. }
    \label{fig:answer-verification-prompt}
\end{figure*}

\section{Human Evaluation} \label{sec:app_human_evaluation}
In this section, we provide comprehensive details regarding our human evaluation protocol.

\subsection{Datasets and Models.}
We evaluate 60 examples, with a random sample of 20 from the three target datasets: \gsm8k, \2wiki, and \scibench. Each example is evaluated independently by three annotators.
We use the Qwen2.5-7B-Instruct model as the thinking LLM across all evaluated systems.

\subsection{Evaluation Criteria.}
Each output is evaluated along four key dimensions. We provide annotators with the following definitions for each factor, which offer additional guidance beyond the brief descriptions in Table~\ref{fig:eval_criteria}.
\begin{itemize}
\item \textbf{Naturalness:} measures whether the response sounds like something a real person would say in a conversation. This is NOT a measure of acoustic quality—focus on the wording and phrasing, not the voice.
\item \textbf{Conciseness:} measures whether the response gets to the point without including unnecessary or excessive information. Focus on whether the response is brief and relevant, or if it feels too long or contains details that aren’t needed.
\item \textbf{Understandability:} measures how clearly the response communicates its meaning. Focus on whether you can easily grasp what the response is trying to say, without getting lost or confused by the way the information is presented.
\item \textbf{Overall Quality:} measures your general impression of the response, taking into account all aspects such as clarity, naturalness, and conciseness. Focus on how well the response works as a whole.
\end{itemize}

\paragraph{Annotation Procedure.}
We recruit the annotators via Amazon Mechanical Turk (MTurk). For each data point, we collect ratings from three independent workers to mitigate subjectivity. We provide the annotators with the following instructions:
\begin{itemize}
\item Carefully read the question and listen to the speech-based response before rating.
\item Rate each evaluation criterion on a 1–5 Likert scale, where 1 represents the lowest and 5 the highest quality.
\item For each criterion, provide a brief explanation to justify the assigned score.

\end{itemize}
Compensation rates are set at \$0.5 per example for \gsm8k and \2wiki, and \$0.7 per example for \scibench, reflecting the varying complexity and required annotation effort across datasets.
Based on the compensation rate per example and average completion time, all participants receive above minimum wage compensation. Participants provide informed consent for the use of their annotations for research purposes.
Explanations are manually reviewed to filter out low-effort or inconsistent responses.

\begin{table*}[tb!]
    \centering
    \scriptsize
    \setlength\tabcolsep{4pt}
    \begin{tabularx}{\textwidth}{lYYYYYYYYYYYY}
    \toprule
        & \multicolumn{3}{c}{WC(↓)} & \multicolumn{3}{c}{FRE(↑)} & \multicolumn{3}{c}{DD(↓)} & \multicolumn{3}{c}{NV(↓)}\\
        \cmidrule(lr){2-4} \cmidrule(lr){5-7} \cmidrule(lr){8-10} \cmidrule(lr){11-13}
        Models & \mhqa & \gsm8k & \scibench & \mhqa & \gsm8k & \scibench  & \mhqa & \gsm8k & \scibench  & \mhqa & \gsm8k & \scibench \\
    \midrule
        \textbf{\textit{Qwen2.5-Omni-7B}} & 49.0 & 101.7 & 138.1 & 74.8 & 90.9 & 72.6 & 5.51 & 5.24 & 5.95 & 0.004 & 0.78 & 30.8 \\
    \midrule
        \multicolumn{13}{l}{\textit{\textbf{Qwen2.5-7B-Instruct}}} \\
        \cmidrule(lr){1-13}
        ~~CoT & 149.4 & 194.5 & 436.0 & 51.0 & 69.9 & 61.6 & 6.51 & 6.42 & 7.18 & 3.737 & 9.079 & 327.2 \\
        ~~SFP & 53.2 & 101.7 & 326.5 & 59.2 & 88.0 & 68.9 & 6.44 & 5.33 & 6.22 & 0.449 & 2.887 & 122.6 \\
        ~~SFF & 40.6 & 48.0 & 80.8 & 70.8 & 88.4 & 75.8 & 4.48 & 4.25 & 5.13 & 0.011 & 0.034 & 5.6 \\
        \cmidrule(lr){1-13}
        ~~\textbf{TVS (\SEQ)} & 46.6 & 43.9 & 90.0 & 71.7 & 88.7 & 74.7 & 4.80 & 4.27 & 5.49 & 0.021 & 0.026 & 5.7 \\
  ~~\textbf{TVS (\model)} & 30.5 & 44.9 & 78.1 & 69.5 & 88.9 & 74.6 & 4.79 & 4.20 & 5.37 & 0.075 & 0.043 & 3.2 \\
    \midrule
        \multicolumn{13}{l}{\textit{\textbf{Llama-3.1-8B-Instruct}}} \\
        \cmidrule(lr){1-13}
        ~~CoT & 145.6 & 153.4 & 296.2 & 58.4 & 69.2 & 61.8 & 5.85 & 6.23 & 7.08 & 14.604 & 67.109 & 384.7 \\
        ~~SFP & 27.9 & 87.1 & 212.4 & 61.8 & 85.0 & 68.4 & 5.70 & 5.45 & 6.69 & 0.155 & 11.043 & 243.0 \\
        ~~SFF & 39.9 & 44.9 & 84.6 & 71.7 & 88.3 & 75.0 & 4.48 & 4.28 & 5.29 & 0.015 & 0.035 & 12.5 \\
        \cmidrule(lr){1-13}
        ~~\textbf{TVS (\SEQ)} & 45.9 & 42.1 & 79.3 & 70.9 & 88.7 & 72.9 & 4.62 & 4.23 & 5.52 & 0.019 & 0.005 & 4.8 \\
  ~~\textbf{TVS (\model)} & 41.6 & 44.0 & 76.9 & 71.1 & 88.4 & 72.8 & 4.55 & 4.21 & 5.43 & 0.173 & 0.024 & 1.9 \\

    \midrule
        \multicolumn{13}{l}{\textit{\textbf{gpt-4o-mini-2024-07-18}}} \\
        \cmidrule(lr){1-13}
        ~~CoT & 104.6 & 175.4 & 291.0 & 53.8 & 67.4 & 63.5 & 5.85 & 6.37 & 6.55 & 11.463 & 74.687 & 352.6 \\
        ~~SFP & 20.4 & 73.1 & 213.8 & 58.4 & 82.5 & 57.3 & 5.79 & 5.14 & 6.37 & 0.011 & 0.215 & 54.0 \\
        \cmidrule(lr){1-13}
        ~~\textbf{TVS (\SEQ)} & 43.1 & 43.8 & 79.4 & 69.7 & 88.5 & 72.8 & 4.63 & 4.27 & 5.52 & 0.014 & 0.008 & 6.4 \\
  ~~\textbf{TVS (\model)} & 32.2 & 45.9 & 74.8 & 70.6 & 88.4 & 73.1 & 4.63 & 4.25 & 5.52 & 0.041 & 0.019 & 2.3 \\

    \bottomrule
    \end{tabularx}
    \caption{
        Detailed speech-suitability scores across all three datasets, comparing various approaches: chain-of-thought (CoT), speech-friendly prompting (SFP), finetuning (SFF), Qwen2.5-Omni-7B, and our proposed methods, TVS (\SEQ) and TVS (\model). Results show consistent trends across all datasets. As we move from \2wiki (denoted as 2MHQA) to \gsm8k and then \scibench, the tasks increasingly demand stronger reasoning capabilities. Correspondingly, the length of test-time reasoning grows and overall speech-suitability decreases.
    }
    \label{tab:app_results}
\end{table*}

\begin{table}
\centering
\begin{tabular}{lrrrrr}
\toprule
SRCC & Natu. & Conc. & Unde. & Over. \\
\midrule
Natu. & 1.000 & 0.757 & 0.550 & 0.878 \\ 
Conc. & 0.757 & 1.000 & 0.594 & 0.761 \\ 
Unde. & 0.550 & 0.594 & 1.000 & 0.700 \\ 
Over. & 0.878 & 0.761 & 0.700 & 1.000 \\
\bottomrule
\end{tabular}
\caption{Inter-criteria correlations for human evaluation metrics using Spearman's rank correlation coefficient (SRCC). The table presents pairwise correlations between Naturalness (Natu.), Conciseness (Conc.), Understandability (Unde.), and Overall Quality (Over.) scores.}
\label{tab:srcc}
\end{table}

\section{Additional Analysis} \label{sec:app_analysis}
In this section, we present additional analyses of our experimental results to complement the main findings discussed in the paper. We provide qualitative examples for each dataset and method, along with detailed dataset-wise results of speech-suitability scores and human evaluation results.

\subsection{Speech-suitability Scores}
We provide speech-suitability scores for individual datasets in Table~\ref{tab:app_results}. We observe that speech-friendly prompting falls short in speech-suitability scores, especially on the \scibench benchmark. We attribute these performance gaps to the \Think model's inability to maintain speech-friendly formatting when addressing complex technical questions that require multiple reasoning steps and computations. In such cases, the model reverts to standard chain-of-thought responses despite explicit instructions for speech adaptation. This finding suggests that prompt-based approaches alone prove insufficient for speech adaptation in highly technical domains.

Additionally, we present  inter-criteria correlation statistics from the human evaluation scores in Table~\ref{tab:srcc}. Notably, Understandability exhibit significantly lower correlation with other criteria, suggesting that comprehensible responses may introduce verbosity and compromise speech naturalness through excessive detail and repetition.

\subsection{Qualitative Results}
To provide a deeper insight into the \framework framework and the \model{} model, we present representative qualitative examples from each evaluation dataset. All examples use Qwen2.5-7B-Instruct as the \Think model. Specifically, Table~\ref{tab:app_qual_gsm8k} presents results on \gsm8k{}, Table~\ref{tab:app_qual_2wiki} on \2wiki{}, and Tables~\ref{tab:app_qual_sci1} and~\ref{tab:app_qual_sci2} on \scibench{}. These examples highlight the strengths and limitations of various approaches in terms of both reasoning capability and speech-friendliness. Our framework demonstrates balanced performance, producing outputs that are logically sound and well-suited for verbal communication.

\begin{table*}
    \centering
    \small
    \begin{tabularx}{\textwidth}{l|X}
        \toprule
        \multicolumn{2}{l}{\textbf{Sample text responses for various methods on \gsm8k}} \\
\midrule
Question & Each class in a school has 20 students. There are 3 classes. The school as a whole is 50\% boys and 50\% girls. The first class has 15 girls. The second class has 12 girls. How many boys are in the third class? \\
\midrule
Answer & 17 \\
\midrule
Chain-of-Thought & To determine the number of boys in the third class, we need to follow these steps:\newline 1. \textbf{Calculate the total number of students in the school:}\newline    - Each class has 20 students.\newline    - There are 3 classes.   $$    \text{Total number of students} = 20 \times 3 = 60   $$ 2. \textbf{Determine the number of boys and girls in the school:}\newline    - The school is 50\% boys and 50\% girls.    $$    \text{Number of boys} = \frac{50}{100} \times 60 = 30   $$    $$    \text{Number of girls} = \frac{50}{100} \times 60 = 30   $$ 3. \textbf{Find the number of girls in the first and second classes:}\newline    - The first class has 15 girls.\newline    - The second class has 12 girls.\newline 4. \textbf{Calculate the total number of girls in the first two classes:}    $$   \text{Total number of girls in first and second classes} = 15 + 12 = 27    $$ 5. \textbf{Determine the number of girls in the third class:}\newline    - Total number of girls in the school is 30.    $$    \text{Number of girls in the third class} = 30 - 27 = 3   $$ 6. \textbf{Calculate the number of boys in the third class:}\newline    - Each class has 20 students.\newline    - The third class has 3 girls.   $$    \text{Number of boys in the third class} = 20 - 3 = 17   $$ Thus, the number of boys in the third class is $\boxed{\correct{17}}$. \\
\midrule
Speech-Friendly Prompting & In each class, there are 20 students. Since the school is 50\% boys and 50\% girls, each class also has 10 boys and 10 girls. The first class has 15 girls, which means it has 5 boys (since it's supposed to have 10 girls, but there are 15, so 20 - 15 = 5 boys). The second class has 12 girls, which means it has 8 boys (since it's supposed to have 10 girls, but there are 12, so 20 - 12 = 8 boys). So far, we have:\newline - First class: 5 boys \quad - Second class: 8 boys\newline That's a total of 13 boys in the first two classes. Since each class is supposed to have 10 boys, the third class must have 7 boys (because 10 - 3 = 7, where 3 is the number of boys already counted in the first two classes). So, there are \wrong{7} boys in the third class. \\
\midrule
Speech-Friendly Finetuning & The first class has 15 girls, so it has 5 boys. The second class has 12 girls, which means it has 8 boys. Together, these two classes have 23 boys. Since each class has 20 students, there are 60 boys in total. That means the third class must have \wrong{37} boys. \\
\midrule
TVS (\SEQ) & The school has 30 boys and 30 girls. The first class has 15 girls. The second class has 12 girls. Together, they have 27 girls. That means the third class has 3 girls. Since each class has 20 students, the third class has 17 boys. So, there are \correct{17} boys in the third class.  \\
\midrule
TVS (\model) & There are 60 students in total. The school is half boys and half girls. That means there are 30 boys. The first class has 15 girls. The second class has 12 girls. Together, they have 27 girls. If we take 27 away from 30, we get 3 girls in the third class. Since each class has 20 students, the third class has \correct{17} boys. \\
\bottomrule
    \end{tabularx}
    \caption{Sample generation results for various methods on \gsm8k. Our models, TVS (\SEQ) and TVS (\model), consistently produce accurate answers with logically sound, step-by-step reasoning, while maintaining high speech-suitableness. In contrast, speech-friendly baseline methods frequently generate answers that are not only incorrect but also logically flawed.}
    \label{tab:app_qual_gsm8k}
\end{table*}

\begin{table*}
    \centering
    \small
    \begin{tabularx}{\textwidth}{l|X}
        \toprule
        \multicolumn{2}{l}{\textbf{Sample text responses for various methods on \2wiki}} \\
\midrule
Question & Which country John Of Bohemia's mother is from? \\
\midrule
Answer & Germany \\
\midrule
Chain-of-Thought & To determine which country John of Bohemia's mother is from, we need to follow these steps:\newline 1. \textbf{Identify John of Bohemia}: John of Bohemia was a Bohemian nobleman and king who reigned as King of Bohemia from 1347 to 1349.\newline 2. \textbf{Research his family background}: John of Bohemia was born in Prague, Bohemia (now part of the Czech Republic) in 1319. His father was Wenceslaus II, Holy Roman Emperor, and his mother was Elisabeth of Bavaria.\newline 3. \textbf{Determine Elisabeth of Bavaria's origin}: Elisabeth of Bavaria was born in 1294 in Landshut, Bavaria, which is now part of Germany. She was a member of the House of Wittelsbach.\newline Given this information, we can conclude that John of Bohemia's mother, Elisabeth of Bavaria, is from Bavaria, which is a region in modern-day Germany.\newline Therefore, the country John of Bohemia's mother is from is \correct{Germany}. \\
\midrule
Speech-Friendly Prompting & John of Bohemia's mother was from France. Specifically, she was from the region of \wrong{Champagne}. \\
\midrule
Speech-Friendly Finetuning & John of Bohemia was a king of Bohemia. His mother was Margaret of Durazzo. She was the daughter of Stephen Dušan, who ruled Serbia. So, John of Bohemia’s mother was from \wrong{Serbia}. \\
\midrule
TVS (\SEQ) & John of Bohemia was a Bohemian nobleman and king. His mother was Elisabeth of Bavaria. She was born in Landshut, Bavaria. That area is now part of Germany. So, John of Bohemia’s mother is from \correct{Germany}.  \\
\midrule
TVS (\model) & John of Bohemia was a Bohemian nobleman and king. He was born in Prague and had a Bavarian mother named Elisabeth. Elisabeth was born in Landshut, Bavaria. So, John of Bohemia's mother is from \correct{Germany}. \\

\bottomrule
    \end{tabularx}
    \caption{Sample generation results for various methods on \2wiki. Compared to baselines, both TVS (\SEQ) and TVS (\model) deliver more accurate reasoning and maintain higher speech-suitableness in their responses. While baseline methods frequently produce incorrect answers or include irrelevant information, our models consistently provide factually correct and well-structured explanations that are both clear and suitable for spoken delivery. This highlights the effectiveness of our approach in balancing reasoning capability with speech-oriented generation quality.}
    \label{tab:app_qual_2wiki}
\end{table*}

\begin{table*}
    \centering
    \small
    \begin{tabularx}{\textwidth}{l|X}
        \toprule
        \multicolumn{2}{l}{\textbf{Sample text responses for various methods on \scibench (1)}} \\
\midrule
Question & Let $X$ and $Y$ be independent with distributions $N(5,16)$ and $N(6,9)$, respectively. Evaluate $P(X>Y)=$ $P(X-Y>0)$. \\
\midrule
Answer & 0.4207 \\
\midrule
Chain-of-Thought & To evaluate $ P(X > Y) $, we need to find the probability that $ X - Y > 0 $. Given that $ X $ and $ Y $ are independent normal random variables, we can determine the distribution of $ X - Y $.\newline \#\#\# Step 1: Determine the Distribution of $ X - Y $\newline Since $ X \sim N(5, 16) $ and $ Y \sim N(6, 9) $:\newline - The mean of $ X $ is $ \mu_X = 5 $.\newline - The variance of $ X $ is $ \sigma_X^2 = 16 $, so the standard deviation is $ \sigma_X = 4 $.\newline - The mean of $ Y $ is $ \mu_Y = 6 $.\newline - The variance of $ Y $ is $ \sigma_Y^2 = 9 $, so the standard deviation is $ \sigma_Y = 3 $.\newline For the difference $ X - Y $:\newline - The mean of $ X - Y $ is:\newline   $$\newline   \mu_{X - Y} = \mu_X - \mu_Y = 5 - 6 = -1\newline   $$\newline - The variance of $ X - Y $ is:\newline   $$\newline   \sigma_{X - Y}^2 = \sigma_X^2 + \sigma_Y^2 = 16 + 9 = 25\newline   $$\newline   Therefore, the standard deviation of $ X - Y $ is:\newline   $$\newline   \sigma_{X - Y} = \sqrt{25} = 5\newline   $$\newline Thus, $ X - Y \sim N(-1, 25) $.\newline \#\#\# Step 2: Standardize $ X - Y $\newline We need to find $ P(X - Y > 0) $. To do this, we standardize $ X - Y $:\newline $$\newline Z = \frac{(X - Y) - \mu_{X - Y}}{\sigma_{X - Y}} = \frac{(X - Y) - (-1)}{5} = \frac{X - Y + 1}{5}\newline $$\newline We want to find:\newline $$\newline P(X - Y > 0) = P\left(\frac{X - Y + 1}{5} > \frac{0 + 1}{5}\right) = P\left(Z > \frac{1}{5}\right)\newline $$\newline \#\#\# Step 3: Use the Standard Normal Distribution\newline The variable $ Z $ follows a standard normal distribution $ N(0, 1) $. We need to find $ P(Z > 0.2) $.\newline Using standard normal distribution tables or a calculator, we find:\newline $$\newline P(Z > 0.2) = 1 - P(Z \leq 0.2)\newline $$\newline From standard normal tables, $ P(Z \leq 0.2) \approx 0.5793 $.\newline Therefore:\newline $$\newline P(Z > 0.2) = 1 - 0.5793 = 0.4207\newline $$\newline \#\#\# Final Answer\newline $$\newline P(X > Y) = P(X - Y > 0) = \boxed{\correct{0.4207}}\newline $$ \\

\bottomrule
    \end{tabularx}
    \caption{Step-by-step reasoning output from the Chain-of-Thought baseline on SciBench. This example illustrates a correct and complete logical progression, resulting in the correct numerical answer.}
    \label{tab:app_qual_sci1}
\end{table*}

\begin{table*}
    \centering
    \small
    \begin{tabularx}{\textwidth}{l|X}
        \toprule
        \multicolumn{2}{l}{\textbf{Sample text responses for various methods on \scibench (2)}} \\
\midrule
Question & Let $X$ and $Y$ be independent with distributions $N(5,16)$ and $N(6,9)$, respectively. Evaluate $P(X>Y)=$ $P(X-Y>0)$. \\
\midrule
Answer & 0.4207 \\
\midrule
Speech-Friendly Prompting & To find $ P(X > Y) $, we need to look at the distribution of $ X - Y $. Since $ X $ and $ Y $ are independent, the difference $ X - Y $ will also follow a normal distribution.\newline First, let's determine the mean and variance of $ X - Y $:\newline - The mean of $ X $ is 5, and the mean of $ Y $ is 6. So, the mean of $ X - Y $ is:\newline   $$\newline   \mu_{X - Y} = \mu_X - \mu_Y = 5 - 6 = -1\newline   $$\newline - The variance of $ X $ is 16, and the variance of $ Y $ is 9. Since $ X $ and $ Y $ are independent, the variance of $ X - Y $ is:\newline   $$\newline   \sigma^2_{X - Y} = \sigma^2_X + \sigma^2_Y = 16 + 9 = 25\newline   $$\newline   Therefore, the standard deviation of $ X - Y $ is:\newline   $$\newline   \sigma_{X - Y} = \sqrt{25} = 5\newline   $$\newline So, $ X - Y $ follows a normal distribution $ N(-1, 25) $.\newline Now, we want to find $ P(X - Y > 0) $. This is equivalent to finding the probability that a normally distributed random variable with mean $-1$ and standard deviation $5$ is greater than $0$.\newline We can standardize this by converting it to a standard normal distribution:\newline $$\newline Z = \frac{(X - Y) - (-1)}{5} = \frac{X - Y + 1}{5}\newline $$\newline We need to find $ P(Z > \frac{0 + 1}{5}) = P(Z > 0.2) $.\newline Using standard normal tables or a calculator, we find that:\newline $$\newline P(Z > 0.2) \approx 0.4207\newline $$\newline Therefore, $ P(X > Y) = P(X - Y > 0) \approx \correct{0.4207} $. \\
\midrule
Speech-Friendly Finetuning & The difference between X and Y, X minus Y, follows a normal distribution. Its mean is 5 minus 6, which is negative 1. Its variance is the sum of the variances of X and Y, so it’s 25. To find P(X greater than Y), we look at P(X minus Y greater than 0). This is the same as finding the probability that a standard normal variable is greater than 1 divided by 5. That value is about 0.3085. So, P(X greater than Y) is approximately \wrong{0.3085}. \\
\midrule
TVS (\SEQ) & The difference between X and Y is X minus Y. The mean of X minus Y is negative one. The variance is 25, so the standard deviation is five. To find P(X minus Y > 0), we standardize X minus Y. This gives us Z equals (X minus Y plus one) divided by five. We want to find P(Z > 0.2). Since Z is a standard normal variable, we use the standard normal distribution. From the table, P(Z > 0.2) is about 0.4207. So, P(X > Y) is \correct{0.4207}.  \\
\midrule
TVS (\model) & The mean of X minus Y is 5 minus 6, which is -1. The variance is the sum of the variances, 16 plus 9, which equals 25. The standard deviation is the square root of 25, which is 5. To find P(X minus Y is greater than 0), we standardize X minus Y. This gives us a standard normal distribution with mean 0 and standard deviation 5. We want to find the probability that this standard normal is greater than 0.2. Using the standard normal distribution, we find that P(X minus Y is greater than 0) is about \correct{0.42}. \\
\bottomrule
\end{tabularx}
\caption{Sample generation results from various methods on \scibench. Our models, TVS (\SEQ) and TVS (\model), consistently produce accurate answers with logically sound and precise numerical reasoning, while maintaining clarity and suitability for spoken delivery. In contrast, baseline methods frequently exhibit logical errors or numerical calculation mistakes. Interestingly, even when prompted for concise and speech-friendly responses, the models tend to generate still structured and verbose outputs on \scibench, due to the inherent complexity of scientific questions.}
\label{tab:app_qual_sci2}
\end{table*}

\end{document}